\documentclass[letterpaper, 10 pt, conference]{ieeeconf}  

\IEEEoverridecommandlockouts                              

\let\labelindent\relax
\usepackage{enumitem}
\usepackage{graphics} 
\usepackage{epsfig} 
\usepackage{mathptmx} 
\usepackage{times} 
\usepackage{amsmath} 
\usepackage{amssymb}  
\usepackage{amsfonts}
\usepackage{amssymb}
\usepackage{booktabs}
\usepackage{graphicx}
\usepackage[bookmarks=true]{hyperref}
\usepackage{multicol}
\usepackage{multirow}
\usepackage{caption,subcaption}
\usepackage{tikz,graphics,float,epsf}
\usepackage{times}
\usepackage{color, colortbl}
\usepackage{xcolor}
\usepackage{cite}
\usepackage{tablefootnote}
\usepackage{arydshln}
\usepackage[flushleft]{threeparttable}
\usepackage{setspace, xspace}

\definecolor{LightCyan}{rgb}{0.88,1,1}
\definecolor{LightYellow}{rgb}{1,1,0.87}

\usepackage{pifont}
\newcommand{\xmark}{\ding{55}}

\definecolor{es-blue}{rgb}{0,0.4,0.8}

\newcommand{\objnav}{\textsc{ObjectNav}\xspace}

\newcommand{\semexp}{SemExp\xspace}
\newcommand{\eeaux}{EE\textsc{Aux}\xspace}
\newcommand{\stubborn}{\textsc{Stubborn}\xspace}

\newcommand{\localmap}{$\mathcal{M}_L$\xspace}
\newcommand{\explorationgoal}{$g_{exp}$\xspace}
\newcommand{\targetobject}{$g_{tgt}$\xspace}

\title{\LARGE \bf
Stubborn: A Strong Baseline for Indoor Object Navigation*
}

\author{Haokuan Luo$^{1}$ and Albert Yue.$^{1}$ and Zhang-Wei Hong$^{1}$ and Pulkit Agrawal$^{1}$ 
\thanks{*This work is primarily funded by Sony Research and in part by the DARPA Machine Common Sense Program and Army Research Office's Grant Number W911NF-21-1-0328.}
\thanks{$^{1}$ All authors are with Improbable AI Lab in the Department of Electrical Engineering and Computer Science at MIT. 
        {\tt\small \{haokuan, ayue, zwhong, pulkitag\}@mit.edu}}%
}

\begin{document}

\maketitle
\thispagestyle{empty}
\pagestyle{empty}

\begin{abstract}
We present a strong baseline that surpasses the performance of previously published methods on the Habitat Challenge task of navigating to a target object in indoor environments. Our method is motivated from primary failure modes of prior state-of-the-art: poor exploration, inaccurate object identification, and agent getting trapped due to imprecise map construction. We make three contributions to mitigate these issues: (i) First, we show that existing map-based methods fail to effectively use semantic clues for exploration. We present a \textit{semantic-agnostic} exploration strategy (called \stubborn) without any learning that surprisingly outperforms prior work. (ii) We propose a strategy for integrating temporal information to improve object identification.
(iii) Lastly, due to inaccurate depth observation the agent often gets trapped in small regions. We develop a multi-scale collision map for obstacle identification that mitigates this issue.

\noindent
\textit{Website:} \href{https://github.com/Improbable-AI/Stubborn}{https://github.com/Improbable-AI/Stubborn}

\end{abstract}

\section{INTRODUCTION}

The ability to efficiently explore and navigate to objects in indoor environments is critical for many robotic applications. The Habitat Object Navigation Challenge~\cite{savva2019habitat} was designed to benchmark indoor object navigation ability of artificial agents. While there are several variants of the Habitat challenge, we are interested in the version where the agent is tasked to navigate to an object instance (e.g., a bed or a refrigerator) in an unseen indoor environment. This task is known as Object Goal Navigation (\objnav). In \objnav the agent only has access to RGBD observations from first person view of the environment and its pose in the global coordinate frame. Successfully solving the task requires the agent to explore the environment to find the target object and move close to it. The accuracy of state-of-the-art method is 24\% and the efficiency measured as \textit{Success weighted by Path Length (SPL)}~\cite{Anderson2018OnEO} is 8.8\%~\cite{savva2019habitat}. 

It is worth contrasting the performance of \textit{object navigation} against \textit{point navigation}, where the agent is tasked to reach target coordinates specified relative to agent's start location (e.g., \textit{Go 3m South and 2m East relative to start}). In \textit{point navigation} the accuracy of state-of-the-art systems is more than $90\%$. Because at every time step the agent is aware of the distance and direction to the goal, efficiently solving the point navigation task requires the agent to remember previously visited locations (i.e., map) and plan a path to the goal. This task is therefore entirely geometric -- the agent can safely ignore scene semantics other than the knowledge of free space and obstacles. 

In \objnav, the agent must explore its environment to discover the goal resulting in additional challenges compared to point navigation: (i) identifying the target object from visual observations; (ii) exploiting scene semantics to speed up exploration (i.e., \textit{semantic exploration}). For instance, if the goal is to reach a  refrigerator, the agent should not explore bathrooms. Similarly, to locate beds the agent should directly navigate to a bedroom instead of exploring other parts of the house. The stark difference in performance between the \textit{object} and \textit{point} navigation tasks indicates that current methods are inadequate at either one or both of \textit{target object identification} and \textit{semantic exploration}.

To understand shortcomings of state-of-the-art methods, we analyzed winners of the 2020 and 2021 \objnav challenge: Goal-Oriented Semantic Policy Agent (\semexp)~\cite{chaplot2020gose} and End-to-End Auxiliary Task Agent (\eeaux)~\cite{ye2021auxiliary} respectively.  Given that image classification systems on the Imagenet dataset have surpassed 90\% accuracy~\cite{dai2021coatnet}, one wouldn't expect object identification to be a major bottleneck in \objnav. However, even 90\% per frame classification accuracy is insufficient. To see why, consider a scenario where the agent requires 100 actions to reach the goal. A \textit{false detection} in any of the first 99 steps will result in failure. Our investigation reveals that \textit{false detection} is the primary failure mode for both \semexp and \eeaux agents.  

We found the second prominent error mode to be failure to explore. While a substantial fraction of mistakes are due to the agent getting \textit{trapped} or going around \textit{loops}, quite surprisingly we found that the \semexp agent is unable to utilize scene semantics to guide exploration (see Fig~\ref{fig:semexp-ablation}). 

Based on these analysis we present an agent called as \stubborn that surpasses the performance of previous winners. \stubborn makes improvements on three fronts:  

\begin{enumerate}[wide, labelwidth=!, labelindent=0pt]

    \item \textbf{Target Object Detection} is improved by accumulating detection scores across frames.
    
    \item \textbf{Exploration} Given that \semexp agent isn't able to utilize semantics for exploration, we developed a much simpler and \textit{semantics-agnostic} exploration method that is both superior and compute/memory efficient.
    
    \item \textbf{Trap Avoidance} via multi-scale representation of collision map. It prevents the agent from getting trapped due to errors in (i) agent's location resulting from discretization\cite{grisetti2007improved,grisettiyz2005improving, mur2015orb}; (ii) inaccurate depth observations\cite{keselman2017intel}; (iii) and artifacts in scans of indoor spaces used in \objnav challenge~\cite{savva2019habitat}. 
\end{enumerate}

Since \stubborn does not use semantics for exploration and yet outperforms prior methods on \objnav challenge, it is a strong \textit{semantic-agnostic} baseline for future works claiming to utilize semantics for exploration.

\begin{table}[]
\centering
\caption{\small Analyzing failure modes of previous winners of the Habitat Object Navigation Challenge~\cite{batra2020objectnav}. Each row reports the percentage of failures attributed to a particular cause along with $95\%$ confidence intervals. The primary cause of failure is false detection followed by exploration related issues of \textit{loop}, \textit{trapped} and \textit{explore.} The last row reports the overall success rate.}
\resizebox{\columnwidth}{!}{%
\begin{threeparttable}
\begin{tabular}{|c|c|c|c|c|}
\hline
\textbf{Failure Mode}   & \multicolumn{1}{c|}{\textbf{\semexp \%}} & \textbf{EEAux \%}             & \textbf{EEAux GT \%}  & \textbf{\stubborn \%}                  \\ \hline
\textbf{False Detection}   & $47.7 \pm 9.0$ & $35.3 \pm 6.3$ & ${1.4 \pm 1.1}$ & $45.1 \pm 8.8$  \\ 
\textbf{Missed Detection}     & $2.7 \pm 2.3$  & $6.3 \pm 3.2 $ & ${1.4 \pm 1.1} $ & $5.3 \pm 3.6$ \rule[-0.9ex]{0pt}{0pt}\\  \cdashline{1-5}
\textbf{Loop}          & $10.8\pm 5.9$ & $14.7 \pm 4.6$ & $22.2 \pm 6.9$ & ${0.9 \pm 0.9}$ \rule{0pt}{2.6ex}\\ 
\textbf{Trapped}       & $10.8 \pm 5.4$ & $12.5 \pm 4.3$ & $13.9 \pm 5.8$ & ${8.0 \pm 5.0}$\\ 
\textbf{Explore}       & $8.1 \pm 5.0  $  & $8.9 \pm 3.7 $ & $12.5 \pm 5.4$ & $13.3 \pm 6.1$ \rule[-0.9ex]{0pt}{0pt}\\  \cdashline{1-5} 
\textbf{Stairs}        & $12.6\pm 5.9$  & $0.4 \pm 0.9$  & $6.9 \pm 4.0$ &$15.9 \pm 6.0$ \rule{0pt}{2.6ex} \\ 
\textbf{Misc}        & $7.2\pm 4.0$  & $21.9 \pm 5.4$  & $41.7 \pm 8.5$ &$11.5 \pm 5.7$ \rule{0pt}{2.6ex} \\ \hline 
\textbf{Success Rate}          &  17.9                           & 23.7  & 40.4\tnote{1} & 23.7\\ \hline 
\end{tabular}
\begin{tablenotes}
    \item[1] Estimated based on results from \cite{ye2021auxiliary} 
  \end{tablenotes}
\end{threeparttable}
}
\label{tab:all_failure}
\vspace{-15pt}
\end{table}

\section{Experimental Setup}
\label{sec:setup}

\objnav task requires the agent to navigate to an instance of the given target object category, such as ``bed'' or ``table'', in an unseen indoor environment. The input to the agent are first person observations from RGBD and GPS+Compass sensors providing location and orientation of the agent relative to its starting position. At each time step, the agent predicts one out of the four actions: \texttt{move forward}, \texttt{turn left}, \texttt{turn right}, and \texttt{stop}. The maximum number of steps in an episode is 500. 

We follow standard evaluation protocols used in the \objnav task of the Habitat Challenge~\cite{chaplot2020gose,ye2021auxiliary}. Performance of various models is evaluated using $2000$ episodes of the validation split of the Matterport 3D dataset~\cite{mp3d}. We test final performance on the Habitat's Challenge evaluation server, the results of which are publically available. Success Rate and SPL are the two primary performance metrics:

\begin{itemize}[wide, labelwidth=!, labelindent=0pt]
    \item \textbf{Success Rate} is the $\%$ of episodes in which the agent stops within a pre-defined distance threshold (1 meter) from an object belonging to the target category. The agent is not required to face the goal when stopping.
    
    \item \textbf{Success weighted by inverse Path Length (SPL)}~\cite{Anderson2018OnEO}: Two agents that take paths of different lengths will have the same success rate. The SPL metric can distinguish between them by measuring agent's efficiency as:
    $$
    \frac{1}{N}\sum_{i=1}^N S_i \frac{l_i}{\max(p_i,l_i)}
    $$
    where $N$ is the number of episodes, $S_i$ is the indicator variable for success, $l_i$ is the shortest path between the agent and the closest object instance belonging to the goal category (provided by an oracle for the purpose of evaluation only), and $p_i$ is the agent's path length.
\end{itemize}

We use the bootstrap method~\cite{efron1985bootstrap} to report the $95\%$ Confidence Intervals for results we collected.


\section{Analyzing Previous Winners}
\label{sec:analysis-previous}
 The winners of \objnav challenge in 2020 and 2021: \semexp~\cite{chaplot2020gose} and \eeaux~\cite{ye2021auxiliary} used substantially different approaches. While \semexp explicitly builds the environment map, \eeaux is an end-to-end architecture using recurrent neural networks for spatial memory. As such these methods are representative of the state-of-the-art in both map-based and map-free algorithms. 

We grouped the failure modes to result from errors in (i) target object detection (top-most group in Table~\ref{tab:all_failure}); (ii) exploration (second group) and other reasons. Our analysis is inspired by categorization of failure modes presented in EEAux~\cite{ye2021auxiliary} and have the following meaning:

\begin{itemize}[wide, labelwidth=!, labelindent=0pt]
    \item \textbf{False Detection} Detecting a wrong object as the goal.
    \item \textbf{Missed Detection} Failures to detect the object even though it was in view.
    \item \textbf{Loop} Poor exploration due to looping over the same locations.
    \item \textbf{Trapped} Repeated collisions with the same or nearby objects cause the agent getting trapped in coverage. Includes when the agent is trapped in spawn.
    \item \textbf{Explore} Generic failures to find the goal although the agent explores new areas efficiently. Includes semantic failures e.g. going outdoors to find a bed.
    \item \textbf{Stairs} The goal is on a different floor and the agent is unable to navigate up/down stairs.
    \item \textbf{Misc.} Other causes of failure irrelevant to this paper such as mesh artifacts, agent randomly quitting the episode, etc.
\end{itemize}

\subsection{Analysis of \semexp Agent}
\label{sec:analysis-semexp}

\begin{figure}[t!]
\centering
    \includegraphics[width=.48\textwidth]{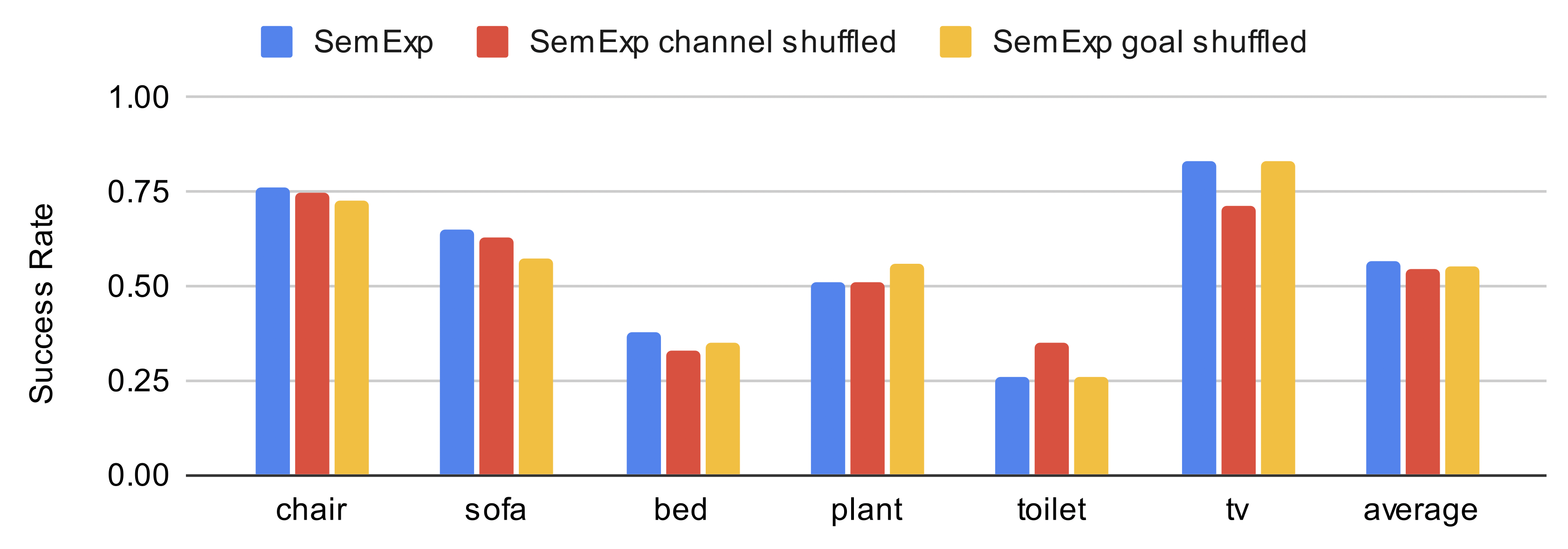}\hfill
    \caption{\small The success rate of \semexp\cite{chaplot2020gose} when using ground-truth semantics (\textit{blue}; $56.5\%$ average accuracy) and two variants where the (i) semantic channels are randomly shuffled (\textit{red}; $54.7\%$ average accuracy) and (ii) the target goal of the planner is set to an incorrect object category (\textit{yellow}; $55.0\%$ average accuracy). The small drop in performance when semantic information is shuffled indicates that \semexp is unable to utilize semantics to guide exploration.} 
    \label{fig:semexp-ablation}
    \vspace{-10pt}
\end{figure}

We analyzed failures of the the \semexp agent by evaluating performance on $200$ validation episodes using pre-trained models released by Chaplot et al. \cite{chaplot2020gose} (see Table~\ref{tab:all_failure}). Since the publically released model only supports $6$ of the $21$ goal categories used in the Habitat challenge, the validation episodes only sample goals from these $6$ object categories. 

Results in Table~\ref{tab:all_failure} show that primary cause of failures are \textit{false detections} accounting for $48\%$ of the mistakes. The next $~\sim 30\%$ of failures are due to exploration challenges of the agent either going around in \textit{loops}, getting \textit{trapped} or simply being unable to find the target object though it was exploring the environment (i.e., \textit{explore}).
To probe the reason behind exploration failures we constructed a version of the \semexp agent using ground-truth semantics. The success rate of this agent was 56.5\%. However, we found that the superior performance is results from mitigation of false detections and not due to better exploration. To understand if the agent is actually using semantics for exploration, we ran two ablation studies where during evaluation we either: (i) shuffled semantic channels or (ii) modified the input of \semexp's global planner that guides exploration to be an object category different from the actual goal. With this change, the agent will attempt to navigate to an incorrect object (e.g., ``toilet" instead of a ``table"). Surprisingly we found that both these ablations only lead to a $2\%$ drop in performance suggesting that \semexp agent is unable to exploit semantics for exploration (see Figure~\ref{fig:semexp-ablation}).


\subsection{Analysis of EEAux Agent~\cite{ye2021auxiliary}}
We evaluated the 6-action model with tethering with the pre-trained weights released by Ye et al. \cite{ye2021auxiliary} over the validation split of the Matterport3D (MP3D) scenes \cite{mp3d}. We collected 300 trajectories with each ground truth and predicted semantics. The dominant failure modes are similar to the \semexp agent. It is worth noting that using ground-truth (GT) semantics results in a much higher success rate indicating that accurate semantic segmentation is a major bottleneck in \objnav. 


\section{Method}
\label{sec:method}

\begin{figure}[t!]
    \includegraphics[width=.48\textwidth]{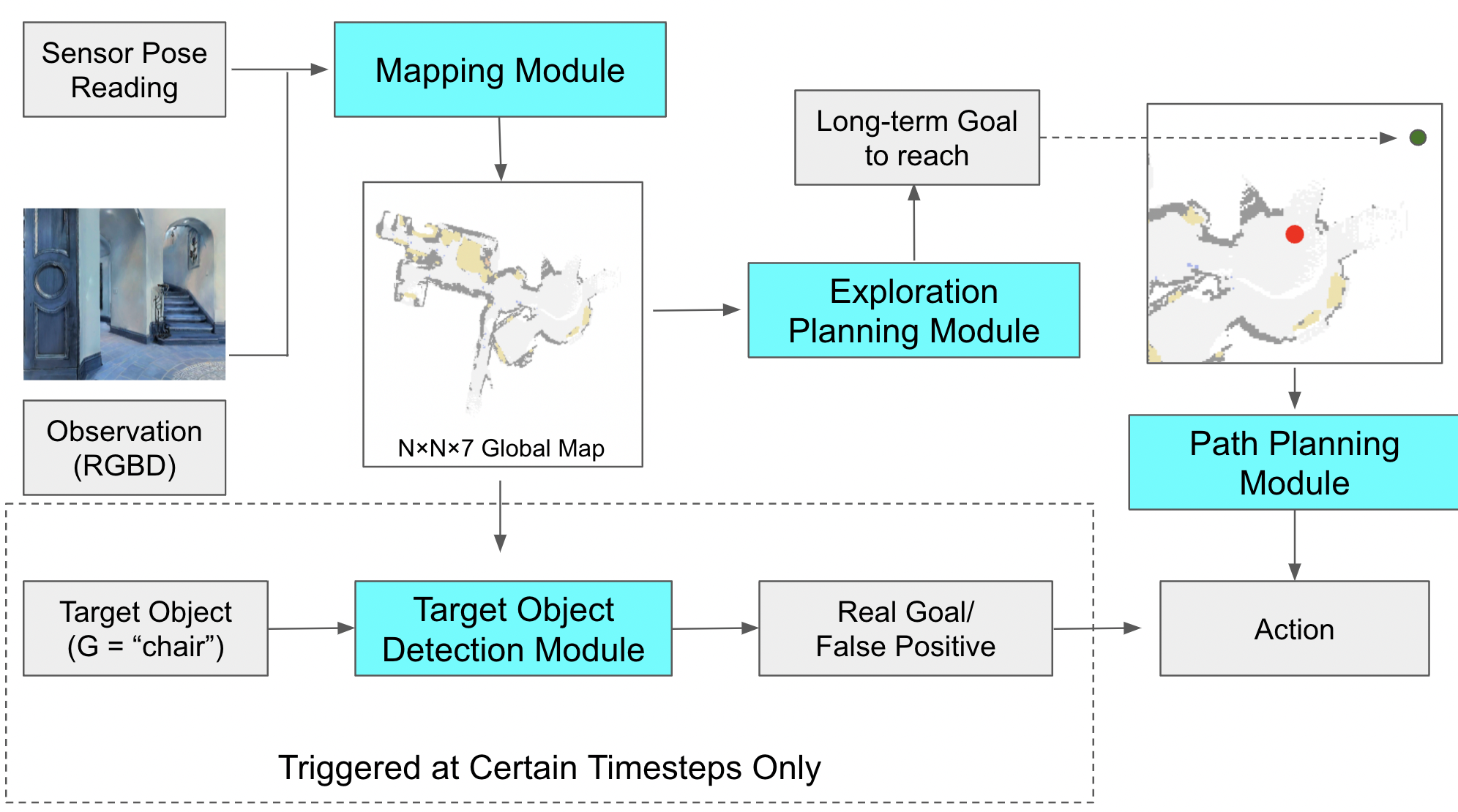}
    \caption{\small Architecture of the \stubborn agent. From sensory observations, the \textit{Mapping} module constructs a top-down 2D map of size $N\times N\times 7$. The \textit{Exploration Planning} module predicts the exploration goal (\explorationgoal; green dot) in a local region around the agent. The \textit{Path Planning} module plans sequence of actions from agent's current location (red dot) to \explorationgoal. The target object detection module }
    \vspace{-5pt}
    \label{fig:overview}
\end{figure}

We propose a modular framework called \stubborn illustrated in Fig. \ref{fig:overview}. \stubborn consists of 
$4$ modules for (i) mapping, (ii) exploration planner, (iii) path planner, and a (iv) multi-frame target object detector. The map is a top-down view of the environment discretized into a 2D grid. Each grid indicates if the space is free or occupied by an obstacle along with other meta-information used by the object detector. The exploration planner takes the map as the input and outputs a 2D grid coordinate indicating the goal for agent's exploration. The path planner outputs a sequence of actions to move the agent from its current location to the goal. The object detector runs at every step of the exploration to determine if the agent has reached the target object. 

\subsection{Mapping Module}
\label{sec:mapping}

\begin{figure}[t!]
\centering
    \includegraphics[width=.47\textwidth]{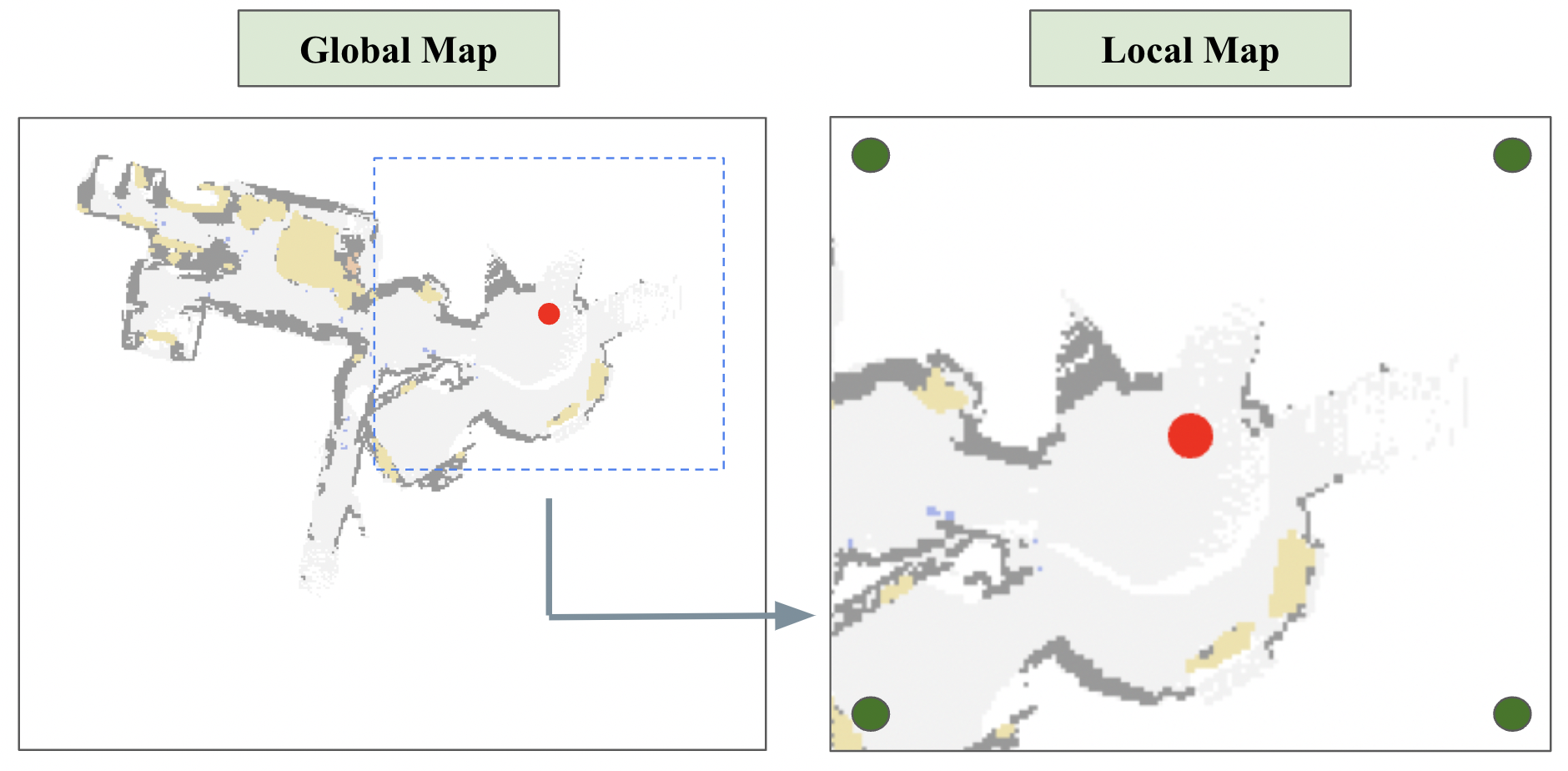}\hfill
    \caption{\small Our agent maintains a global map of size $N\times N$, but plans path using a local map of size $K\times K$ around the agent (the red circle) to save computation time. The four green dots on the corners of the local map are the four possible exploration goal locations.}
    \vspace{-5pt}
    \label{fig:localmap}
\end{figure}

\begin{table}[t!]
\caption{\small A description of the 7-channel map we maintain. The first three channels of the map are used to record obstacle-related information, while the rest of the channels are used to record goal-object-related information.}
\resizebox{\columnwidth}{!}{%
\begin{tabular}{|c|c|l|}
\hline
                                                                                                  & \textbf{Channel} & \multicolumn{1}{c|}{\textbf{Description}}            \\ \hline
\multirow{3}{*}{\textbf{\begin{tabular}[c]{@{}c@{}}Obstacle \\ Map\end{tabular}}}                 & 1                & Obstacle Map Using Depth Input                       \\ \cline{2-3} 
                                                                                                  & 2                & Pessimistic Obstacle Map using Collision Information \\ \cline{2-3} 
                                                                                                  & 3                & Optimistic Obstacle Map using Collision Information  \\ \hline
\multirow{4}{*}{\textbf{\begin{tabular}[c]{@{}c@{}}Object \\ Identification \\ Map\end{tabular}}} & 4                & Total Number of Frames Appeared in View              \\ \cline{2-3} 
                                                                                                  & 5                & Sum of Confidence Score of Target Object               \\ \cline{2-3} 
                                                                                                  & 6                & Maximum Confidence Score of Target Object              \\ \cline{2-3} 
                                                                                                  & 7                & Maximum Confidence Score of Non-Target Object          \\ \hline
\end{tabular}
}
\label{tab:map-channel-description}
\end{table}

The agent maintains a map $\mathcal{M}$ of size $N\times N \times 7$, where $N\times N$ ($7200 \times 7200$) denotes the number of mapping grid units each of size $25cm^2$. The first three out of seven channels are binary, where $1$ indicates obstacles and $0$ represents free space. Unexplored space is treated as free space. Following the method proposed in \semexp, the first channel stores occupancy computed using depth from a RGBD camera. 

Due to several reasons such as inaccurate depth observations, mesh artifacts in the Habitat simulator producing invisible obstacles, just relying on depth data to identify free space is insufficient. To mitigate these concerns, we additionally use agent's collisions to identify obstacles. A collision is defined as an event where the agent attempts to move forward but its location remains unchanged. Unfortunately, we cannot directly use collisions to construct an occupancy map because the agent's location is discretized and the the exact location and size of the obstacle is unobserved. If an obstacle is marked to occupy a size larger than its actual size, it will prevent the planner from planning a path. On the flip side, marking the obstacle to be smaller than actual size will yield infeasible plans. 

We tackle this problem by maintaining a \textit{multi-scale} obstacle map: a \textit{pessimistic map} that marks larger areas ($25\times 25~cm^2$) and an \textit{optimistic map} that marks smaller areas ($15\times 15~cm^2$) in front of the agent as obstacles. Given uncertainty in obstacle size and location, a useful rule of thumb we found was to always mark coordinates along the \textbf{visited path} of the agent as free space. It reduces the chances of an agent getting trapped due to incorrectly marked obstacles. These two collisions maps are encoded in channels 2 and 3 of $\mathcal{M}$ and are used by the path planner (see Sec.~\ref{sec:path-planner}). Channels 4-7 of $\mathcal{M}$ (see Sec.~\ref{sec:object-detection-module}).


\begin{figure*}[t!]
    \includegraphics[width=.24\textwidth]{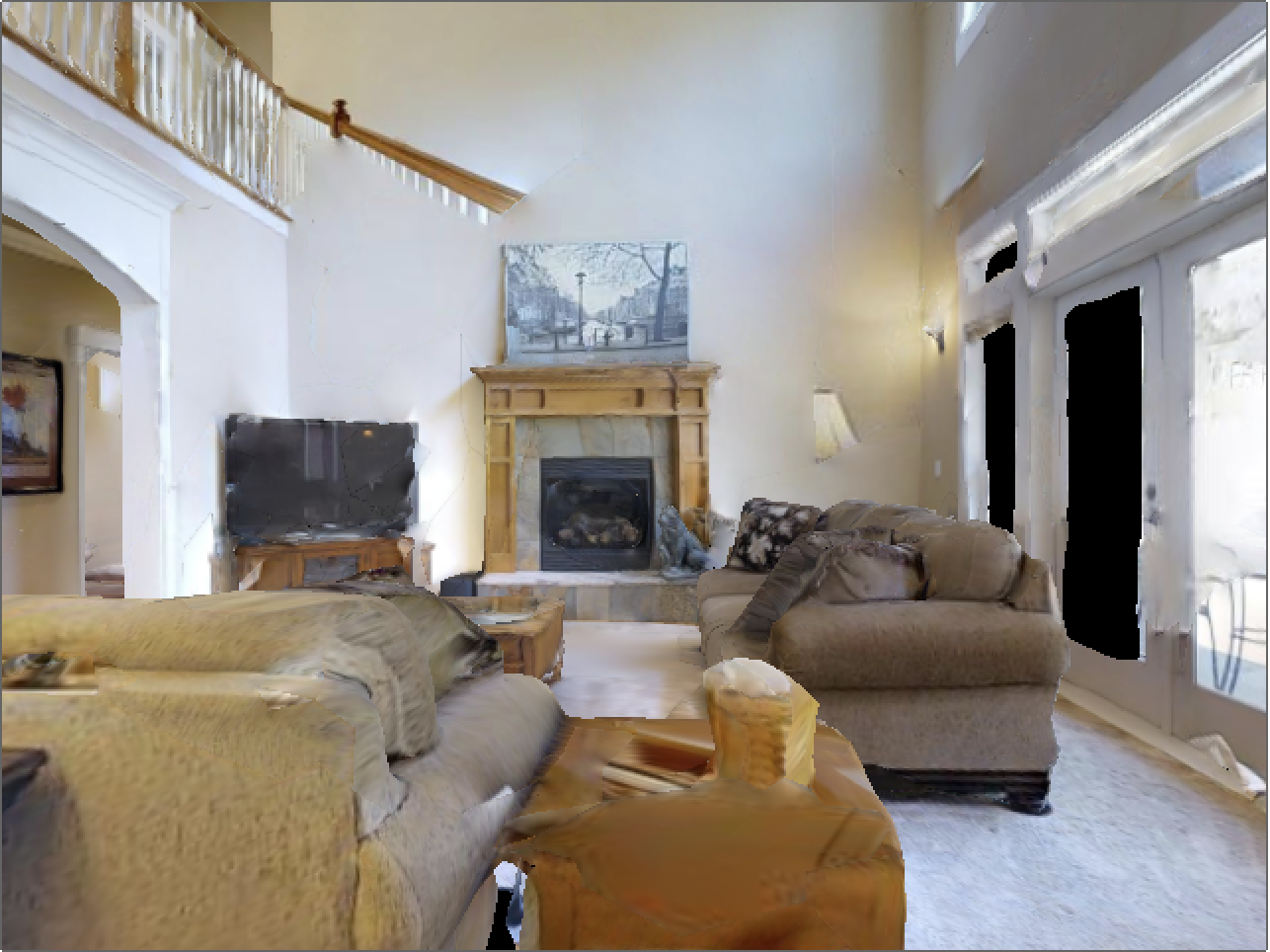}\hfill
    \includegraphics[width=.24\textwidth]{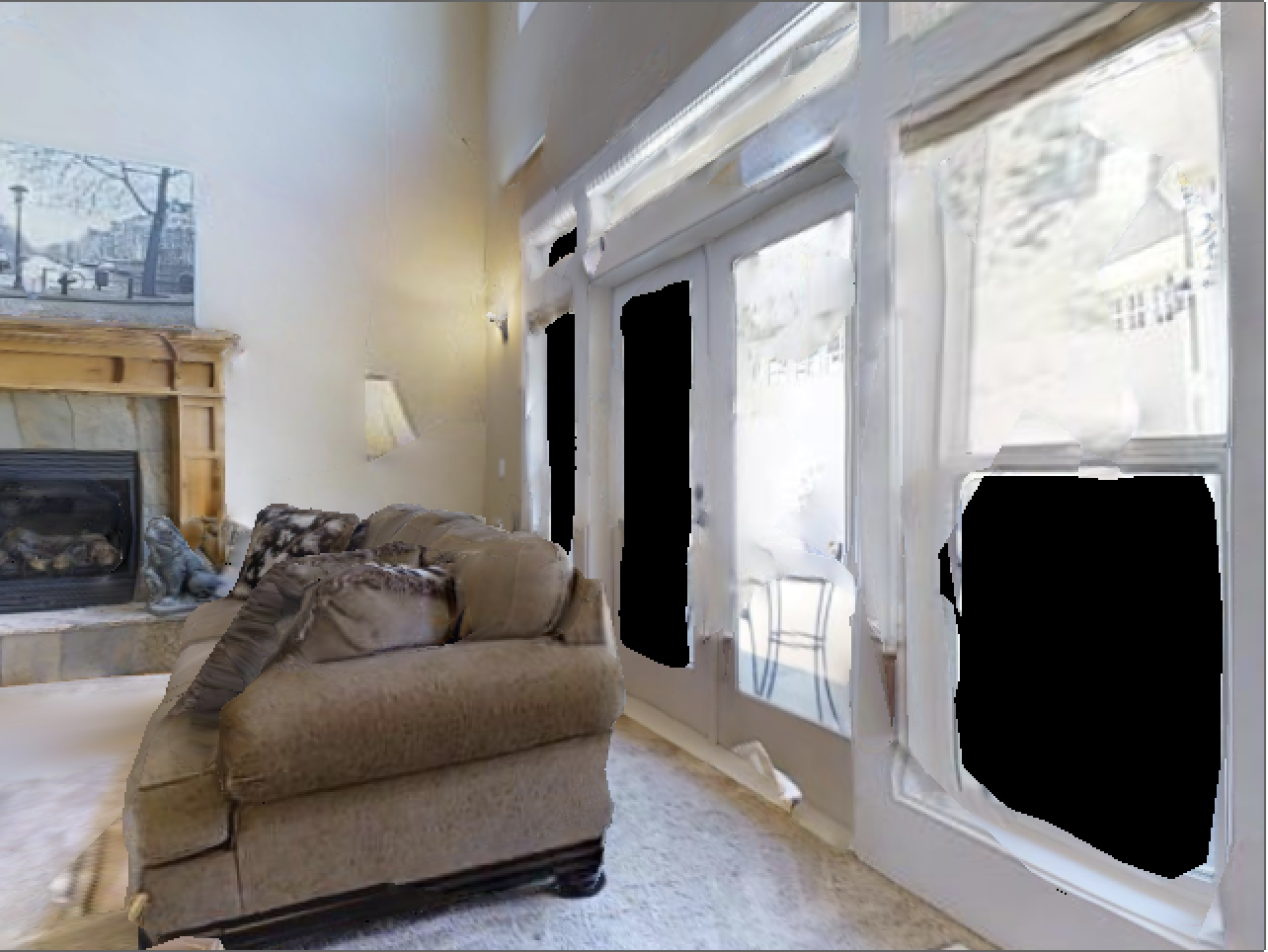}\hfill
    \includegraphics[width=.24\textwidth]{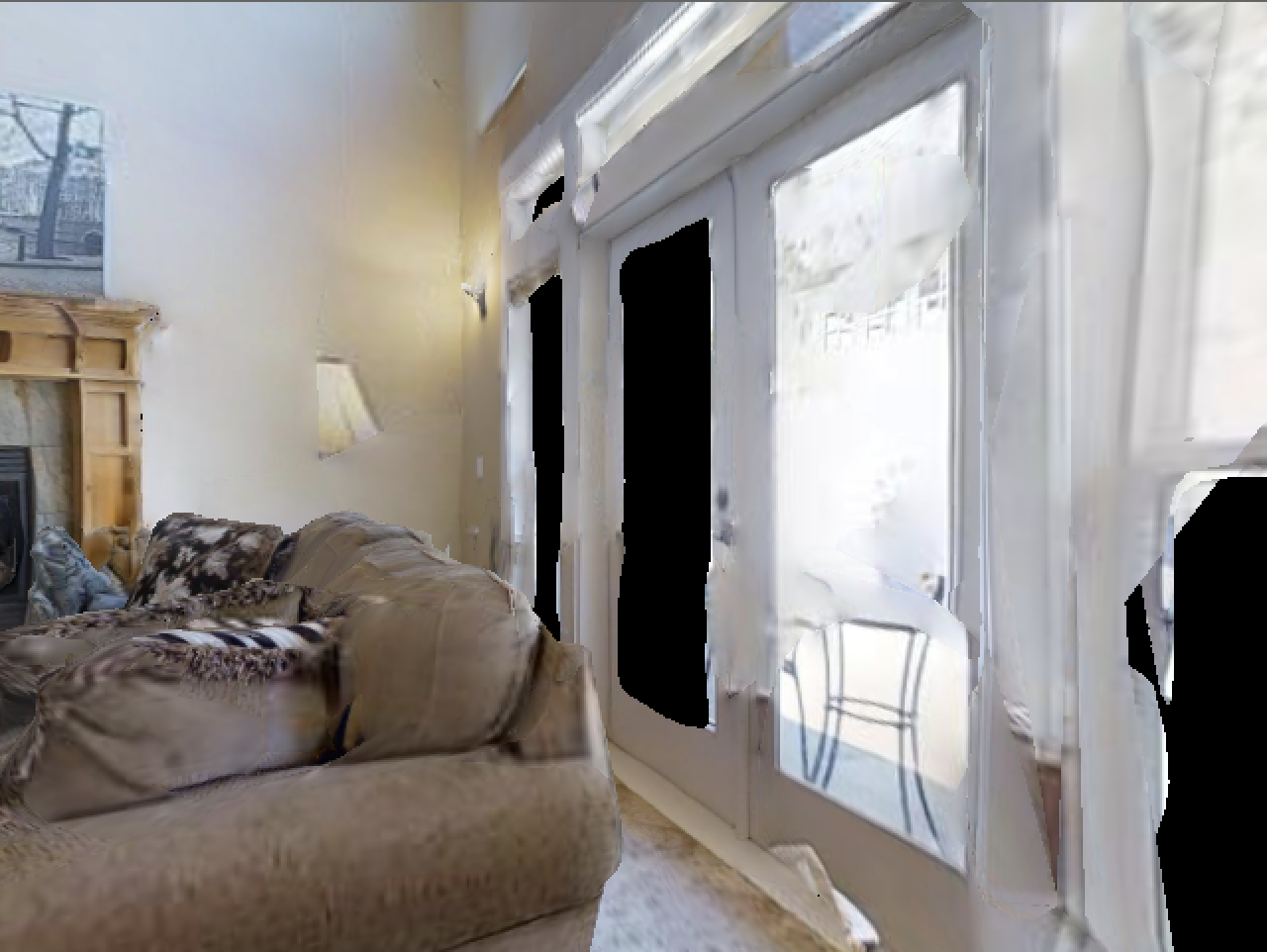}\hfill
    \includegraphics[width=.24\textwidth]{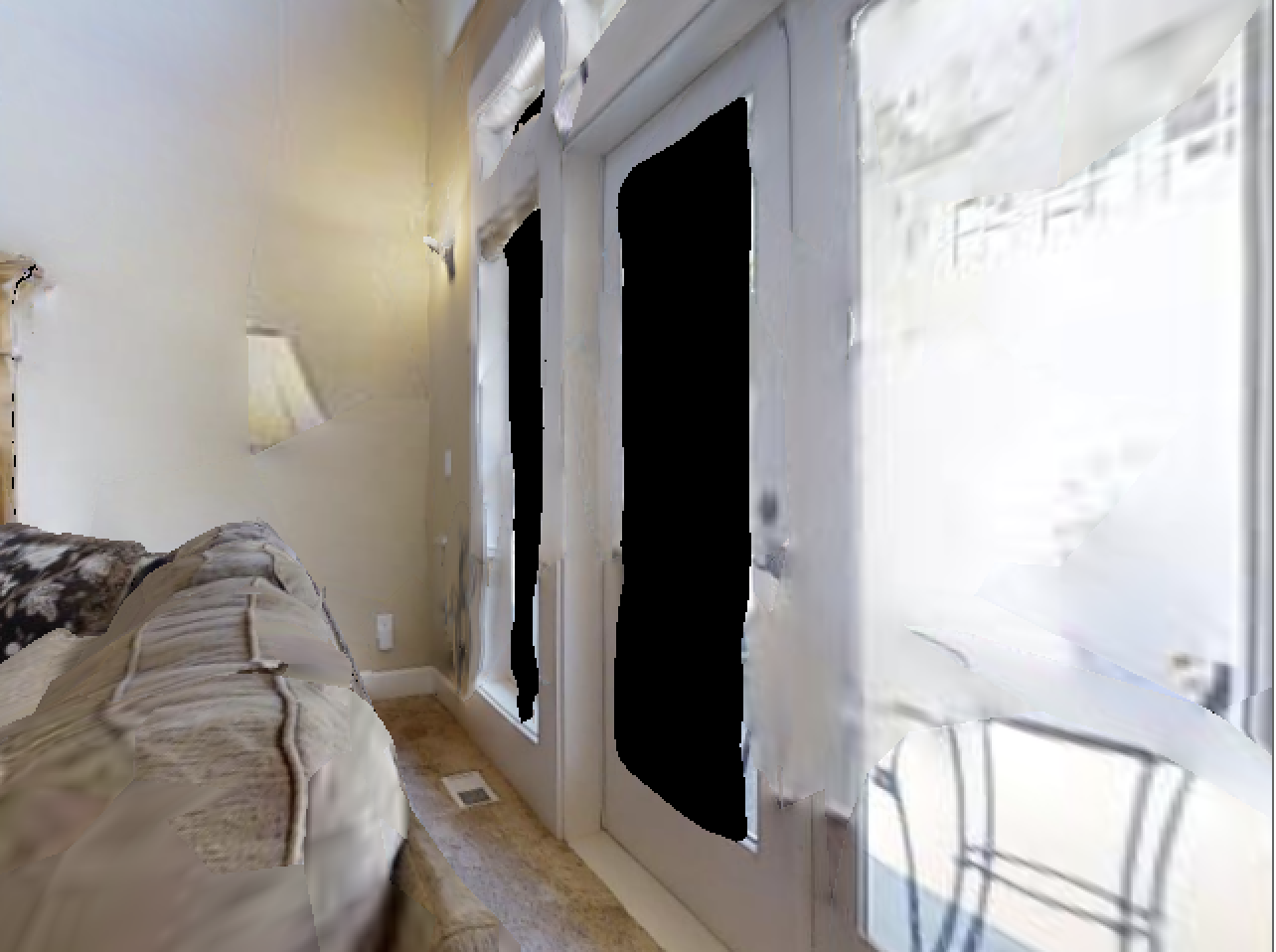}
    \caption{\small Tasked to go to a bed, the agent mistakes the sofa as a bed in the last frame and stops. Such a false detection results in failure.}\label{fig:foobar}
    \label{fig:failure_sofa}
    \vspace{-10pt}
\end{figure*}

\begin{figure}[t!]
\centering
    \includegraphics[width=.45\textwidth]{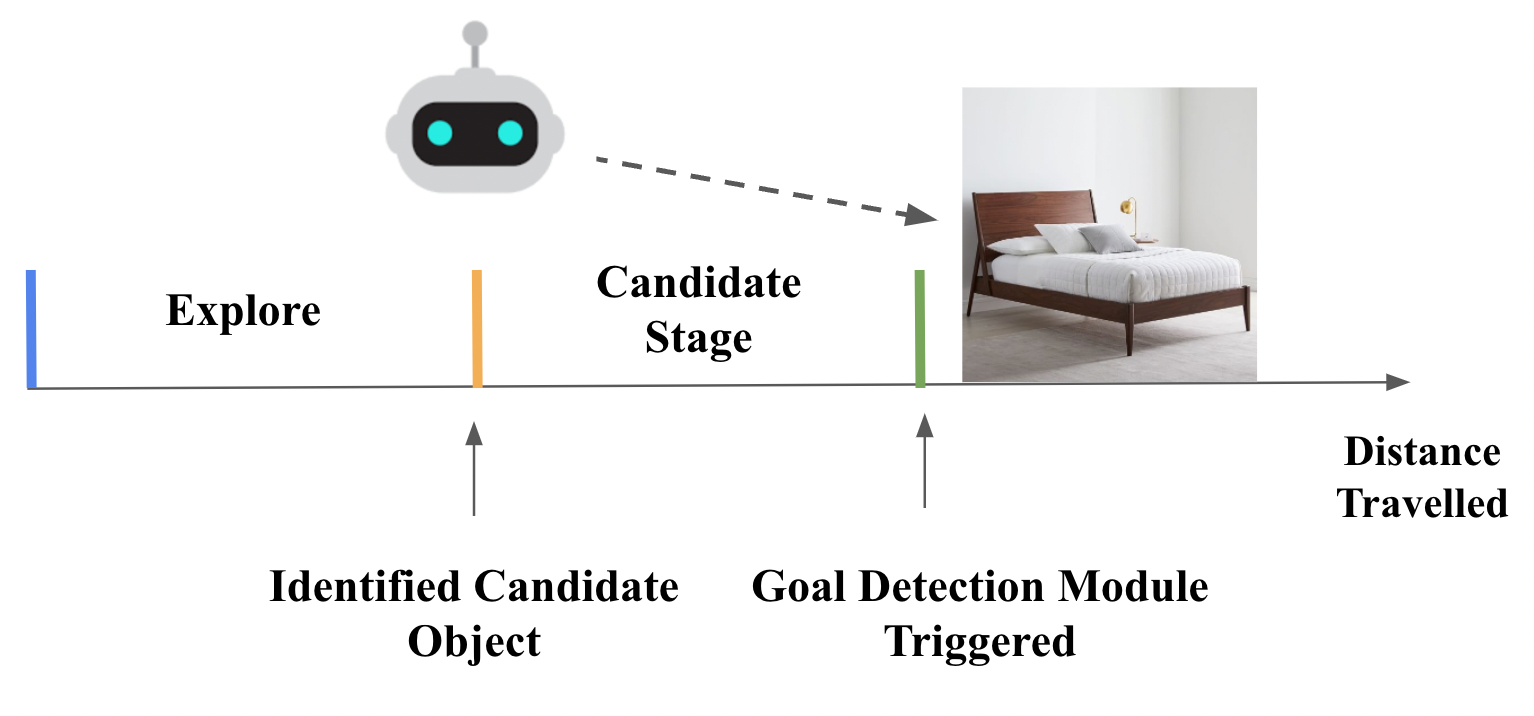}\hfill
    \caption{\small Two stages of Goal Detection. After a candidate object is detected, the agent enters the candidate approach stage. After reaching the candidate object, the agent triggers the target detection module to decide whether to stop or continue exploring.}
    \vspace{-5pt}
    \label{fig:stage_goal_detection}
\end{figure}

\subsection{Exploration Planning Module}
\label{sec:exploration-module}

For discovering the target object the agent must explore its environment. Given we found that some state-of-the-art systems are unable to utilize semantics for exploration (Sec.~\ref{sec:analysis-previous}), we constructed a simple rule-based method for exploration that does not rely on any semantic information. Our intuition is that the agent will discover large areas of the environment without wasteful to-and-fro between two locations if it keeps moving along a particular direction until it runs into a dead end. At the dead-end, the agent can pick another direction and repeat the process.  

Given the first three channels of the map as input ($\mathcal{M}[:,:,1:3]$), the \textit{exploration planner} outputs one out of the four corners of a region of size $K \times K$ (\textit{local map}; \localmap) centered around the agent as the exploration goal (\explorationgoal). Same as in \semexp, we choose $K = 1200$ and the exploration goal is predicted in a local map instead of the global map to speedup computation. The agent is ``\textbf{stubborn}" in the sense that it continues exploring until the \textit{path planning module} (Sec.~\ref{sec:path-planner}) can find no feasible path to the current goal. When such a situation occurs, the agent chooses the next corner in a clockwise direction as its exploration goal.  


\noindent
\textbf{Collision Averse Exploration} In order to pursue the exploration goal, the path planner uses the depth obstacle map and the pessimistic collision map (see Sec.~\ref{sec:mapping}). The use of pessimistic map keeps the agent away from obstacles and thereby reduces the chances of the agent getting trapped (see results in Table~\ref{tab:stubborn-ablation}) and prematurely endings its exploration. 


\subsection{Path Planning Module}
\label{sec:path-planner}

Given the local map of obstacles (\localmap\unskip$[:,:,1:3]$) and grid coordinates of a goal ($g$), following the work of \semexp, we use fast-marching method\cite{ffm} to compute the shortest path from agent's current location to $g$. The path is a sequence of 2D coordinates and subsequent coordinates are used infer one out of the four agent actions. The path is recomputed after every action (i.e., closed loop control). 


\noindent
\textbf{Choice of Obstacle Map} The path planner uses obstacle map that is computed by the element-wise OR operation between the depth obstacle map and the \textit{pessimistic} collision obstacle map $\big($\localmap\unskip$[:,:,1]$ $\lor$ \localmap\unskip$[:,:,2]\big)$. While the path planned using the pessimistic map maybe sub-optimal, it significantly decreases agent's chance of getting stuck and thereby improving the exploration efficiency.

If goal of the planner is \explorationgoal, the pessimistic map is always used. However, if the planner's goal is the location of target object, then if the pessimistic planner leads to no feasible plan, the path is planned using the optimistic collision obstacle map $\big($\localmap\unskip$[:,:,1]$ $\lor$ \localmap\unskip$[:,:,3]\big)$.  

\noindent
\textbf{Brute Force Untrap Mode} Sometimes the free space is so small that the path planner fails to find any feasible plan due to  discretization of the agent's location to grid elements. If the agent fails to move forward (i.e., it is stuck) for ten consecutive steps, the path planner enters a \textit{brute force untrap mode}: the agent turns either left ($L$) or right ($R)$ and then attempts to move forward ($F$). Suppose it chose, left at the first step, then the executed actions are $L,F,L,F,L,F,L,F...$, until one of the forward action results in a forward motion. In such a scenario, the untrap mode ends and the agent resumes normal operation. However, if the agent never moves forward, it stays in the untrap mode until the end of the episode. The choice of left or right motion for untrapping alternates each time the agent enters the untrap mode. 

\subsection{Target Object Detection Module}
\label{sec:object-detection-module}
Prior approaches to target object detection such as the one used by \semexp rely on single-frame confidence score. Because the agent decides at every step whether it reached the target object or not, relying on per-frame decisions leads to a high false detection rate as discussed in Sec.~\ref{sec:analysis-semexp}. E.g., consider the sequence of four frames in Fig.~\ref{fig:failure_sofa} encountered by the agent trying to reach a ``bed". The agent successfully identifies the sofa in the first three pictures, but mistakes the sofa as a bed in the fourth frame. Consequently the agent stops, believing it reached its target, but in reality it has failed. Even to the human eye, in the fourth frame the sofa looks a little like a bed. However, humans won't mistake the sofa as a bed due to the memory of ``sofa" from previous frames~\cite{zhu2012human}. 

To mitigate false detection we designed a \textit{two-stage approach} that detects the target object based on \textit{sequence of frames}. Two stages illustrated in Fig.~\ref{fig:stage_goal_detection} are: (a) Similar to \semexp, the first stage predicts for every grid location in \localmap visible in the agent's view ($V$) the confidence score $\big(c_l \in [0,1]; l \in [1,L]\big)$ for $L$ object categories: $\big\{(x,y) \rightarrow [c_1^{x,y}, c_2^{x,y}, ..., c_L^{x,y}] \forall (x,y) \in V \big\}$. Following \eeaux, we use a RedNet \cite{jiang2018rednet} model finetuned on 100K randomly sampled views from Matterport 3D \cite{ye2021auxiliary} for these predictions. If the presence of target object is detected at a particular grid location $\big($\targetobject: $(x,y)\, \text{s.t.}\, c_{tgt}^{x,y} > 0.5\big)$, then agent starts approaching the detected target object by setting the target object location as the output of the exploration planning module. When the agent is within a distance threshold (1 meter) to the target object, it enters the \textit{verification} stage to reduce the chances of a false detection. 


\noindent
\textbf{Verification Stage} is a binary classifier using a manually designed 4D feature representation that captures temporal information about identity of the target object to decide if its a false detection. The 4D feature vector is stored as 4 channels of the map that the agent builds as it explores the environment. Concretely, given \localmap\unskip$[$\targetobject\unskip$(y)$,\targetobject\unskip$(x),4:7$], the binary classifier outputs 0/1 to indicate absence/presence of target object. \localmap\unskip$[:,:,4:7$] has the following information:
\begin{itemize}[wide, labelwidth=!, labelindent=0pt]
    \item \textbf{Total View(\localmap\unskip$[y,x,4]$):} Number of frames in the episode that \localmap\unskip$[y,x,4]$ appeared in agent's view. 
    
    \item \textbf{Cumulative Confidence (\localmap\unskip$[y,x,5]$):} cumulative confidence scores of target object at location $(x,y)$ until current time step ($t$) computed as $\sum_{i=1}^t c_{tgt}^{x,y}(i)$. 
    
    \item \textbf{Max. Confidence (\localmap\unskip$[y,x,6]$):} $\max_{i \in [1,t]} c_{tgt}^{x,y}(i) $
    
    \item \textbf{Max. Other Confidence(\localmap\unskip$[y,x,7]$):} maximum score among non-target categories, $\max_{l \neq tgt}\big[\max_{i \in [1,t]} c_{l}^{x,y}(i)\big]$
\end{itemize}


\begin{table}[t!]
\centering
\caption{\small The four methods used to avoid collisions help the \stubborn Agent reduces the trap rate and improves exploration efficiency measured by the GT exploration rate. The full definition of abbreviated method names can be found in Sec.~\ref{ssec:collision-avoidance}.}
\resizebox{\columnwidth}{!}{%
\begin{tabular}{|l|cccc|cc|}
\hline
               & \multicolumn{4}{c|}{\textbf{Collision Avoidance Update}}                   & \multicolumn{2}{c|}{\textbf{Performance}}            \\ \hline
\textbf{Agent} & \textbf{BFUntrap} & \textbf{ColObs} & \textbf{ColAver} & \textbf{VisPath} & \textbf{Trapped\%} & \textbf{GT Exploration Rate \%} \\ \hline
\textbf{I}     &                   &                  &                  &                  & 10.0                 & 54                         \\ \hline
\textbf{II}    & \checkmark        &                  &                  &                  & 8.7                 & 59                         \\ \hline
\textbf{III}   & \checkmark        & \checkmark       &                  &                  & 6.7                 & 64                         \\ \hline
\textbf{IV}    & \checkmark        & \checkmark       & \checkmark       &                  & 5.2                 & 64                         \\ \hline
\textbf{V}     & \checkmark        & \checkmark       & \checkmark       & \checkmark       & \textbf{5.0}        & \textbf{67}                \\ \hline
\end{tabular}
}
\label{tab:stubborn-ablation}
\end{table}

The verification system is a binary Naive Bayes Classifier trained using data collected by running the agent on the validation episodes of Matterport 3D to prevent overfitting the classifier to the training episodes which were used to train the base object segmentation model.  A separate classifier is trained for each target object category using trajectories generated by the agent in pursuit of the same object category. At test-time, if the output of classifier is 1, agent believes it has reached the target and stops. Otherwise, the agent marks \targetobject under consideration as a false positive to prevent itself from being tricked by the same location in future exploration. If the first stage of object detection finds the target after 200 time steps (which we choose as a hyperparameter), the verification system is not used. Instead the agent is stopped once it reaches the detected target. The rationale is that close to the end of an episode there is not enough time to search other locations for the target. Therefore, the agent might as well navigate to the detected target location.

\section{Results}
\label{sec:results}

Our main contributions are: (i) semantic-agnostic exploration method that is competitive with state-of-the-art; (ii) multi-scale collision maps for improving exploration and (iii) multi-frame object detection system to reduce false positives. The empirical evaluation presented in subsequent sections supports the importance of these contributions. 

\subsection{Exploration Efficiency}
\label{sec:exploration-efficiency}

\begin{table}[t!]
\caption{\small Comparing the exploration performance of Frontier, \semexp, and \stubborn agents measured as the GT Exploration Rate (Sec.~\ref{sec:exploration-efficiency}) with $95\%$ Confidence Intervals. \semexp and \stubborn are both tested with and without the Collision Avoidance Update described in Sec. \ref{ssec:collision-avoidance}}
\centering
\resizebox{\columnwidth}{!}{%
\begin{tabular}{|c|c|c|c|c|c|c|c|c|}
\hline
\textbf{Methods}          & \textbf{Collision Avoidance Update} & \textbf{chair } & \textbf{sofa} & \textbf{bed}  & \textbf{plant} & \textbf{toilet} & \textbf{tv}               & \textbf{average \%}         \\ \hline
\textbf{Frontier}         & \checkmark                          & 75            & 65          & 24          & 56           & 29            & \textbf{100} & $58 \pm 6$          \\ \hline
\multirow{2}{*}{\textbf{\semexp}}   & \xmark                              & 76            & 65          & 38          & 51           & 26            & 83                      & $57\pm 7$           \\
                          & \checkmark                          & 79            & 65          & 38          & 57           & 26            & 83                      & $58\pm 6$           \\ \hline
\multirow{2}{*}{\textbf{\stubborn}} & \xmark                              & 74            & 58          & 38          & 50            & 36            & 66                      & $54 \pm 6$          \\
                          & \checkmark                          & \textbf{80}    & \textbf{72} & \textbf{39} & \textbf{61}  & \textbf{67}   & 83                      & \textbf{67} $\pm 6$ \\ \hline
\end{tabular}
}

\label{tab:performance3}
\end{table}

The overall success rate of the agent entangles the performance of target detection and exploration. To study exploration efficiency in isolation, we use the following metrics:  

\begin{itemize} [wide, labelwidth=!, labelindent=0pt]
    \item \textbf{Ground Truth (GT) Exploration Rate} To remove the effect of target detection in evaluating exploration performance, we consider the agent to be successful if target object occupies at least $0.8\%$ of the pixels in the \textit{ground truth semantic mask} in the agent's view.
    
    \item \textbf{Trapped Rate} is the percentage of episodes the agent gets trapped. If the agent moves less than $1$ meter in $100$ time steps it is considered trapped. Lower trapped rate is better.
\end{itemize}

We compare exploration performance of \stubborn against \semexp~\cite{chaplot2020gose} and the well known \textit{Frontier Based Exploration} baseline~\cite{frontier} where the agent explores by navigating to the frontier of previously visited locations. We used the publicly released pre-trained model of \semexp. The global map size in the \semexp model is $2400\times 2400$, which we found to be small for Matterport 3D dataset and results in poor exploration. We modified the global map size to $7200\times 7200$ to match with \stubborn. Since public version of \semexp only supports $6$ semantic categories, we evaluate the performance on these categories, with and without the \textit{Collision Avoidance} method mentioned below. 

\subsection{Collision Avoidance Method}
\label{ssec:collision-avoidance}

The \stubborn Agent used $4$ methods to prevent agent from getting trapped: (i) Brute Force Untrap (BFUntrap; Sec.~\ref{sec:path-planner}); (ii) Multi-Scale collision obstacle maps ({ColObs}; Sec.~\ref{sec:path-planner}); (iii) collision aversion in exploration planner (ColAver; Sec.~\ref{sec:exploration-module}); and (iv) marking visited paths as free space (VisPath; Sec.~\ref{sec:mapping}). Results in Table~\ref{tab:stubborn-ablation} show that all four factors are essential for improving exploration and reducing trap frequency. The most significant performance gain result from multi-scale collisions representation (ColObs).

Results in Table \ref{tab:performance3} show that without collision avoidance, \semexp and \stubborn achieve comparable exploration performance. With collision avoidance, \semexp is no better than semantics-agnostic \textit{Frontier} exploration strategy. \stubborn with collision avoidance outperforms all baselines by a large margin. One reason why \stubborn benefits from collision avoidance, but \semexp doesnot is that the \stubborn agent is prone to getting trapped. This is because its exploration goals are chosen to be corners of a local map which forces the agent to go to the corners of the rooms more often, where it is more likely to be trapped. Collision avoidance reduces the chances of the agent getting trapped and therefore benefits \stubborn. Note that the reported results are using data from the validation set and experiments are run with no constraint on execution time. However, for the actual test on the Habitat Challenge evaluation server, the total execution time is constrained.



\subsection{Target Object Identification}
\label{OIR}
\noindent


\noindent
\textbf{Baseline} We compare \stubborn with a single frame baseline (Single-Frame) that only uses the first stage of the target object detection described in Sec.~\ref{sec:object-detection-module}. This baseline is akin to the kind of object detection used in \semexp. 

\noindent
\textbf{Results} Table~\ref{tab:ObjIDResults} compares the performance of the proposed multi-frame object detector (\stubborn) against the single frame baseline on $15$ object categories. We do not report performance on $6$ out of the $21$ categories because these target object categories were contained in $4$ or less house environments. Results show that the proposed object detection scheme improves performance on most object categories. While the average performance of \stubborn is higher than the Single-Frame agent, it is not statistically significant measured by 95\% confidence intervals. At the same time, we consistently noticed performance improvements from the multi-frame object detector. On the official test set of Habitat Challenge, the agent's success rate increased from $22.2\%$ to $23.7\%$, by switching from Single-frame to \stubborn. 

\begin{table}[]
\caption{\small Comparing success Rate of \stubborn against the baseline Single-Frame Agent on $15$ semantic categories, with $95\%$ Confidence Intervals.}
\centering
\resizebox{\columnwidth}{!}{%
\begin{tabular}{|c|c|c|c|}
\hline
\textbf{Goal}               & \textbf{\stubborn \%} & \textbf{Single-Frame \%}  & \textbf{Improvement \%} \\ \hline
\rowcolor[HTML]{E4F8E1} 
\textbf{stool}              & \textbf{16}     & 9            & 77.78                   \\ \hline
\rowcolor[HTML]{E4F8E1} 
\textbf{toilet}             & \textbf{43}     & 35            & 22.86                   \\ \hline
\rowcolor[HTML]{E4F8E1} 
\textbf{sink}               & \textbf{31}     & 26            & 19.23                   \\ \hline
\rowcolor[HTML]{E4F8E1} 
\textbf{chest\_of\_drawers} & \textbf{19}     & 16            & 18.75                   \\ \hline
\rowcolor[HTML]{E4F8E1} 
\textbf{plant}              & \textbf{33}     & 28            & 17.86                   \\ \hline
\rowcolor[HTML]{E4F8E1} 
\textbf{bed}                & \textbf{51}     & 44            & 15.91                   \\ \hline
\rowcolor[HTML]{E4F8E1} 
\textbf{cabinet}            & \textbf{34}     & 30             & 13.33                   \\ \hline
\rowcolor[HTML]{E4F8E1} 
\textbf{table}              & \textbf{61}     & 56            & 8.93                    \\ \hline
\rowcolor[HTML]{E4F8E1} 
\textbf{counter}            & \textbf{39}     & 36            & 8.33                    \\ \hline
\rowcolor[HTML]{E4F8E1} 
\textbf{chair}              & \textbf{47}     & 46            & 2.17                    \\ \hline
\rowcolor[HTML]{F6F7D9} 
\textbf{sofa}               & 31              & 31            & 0                       \\ \hline
\rowcolor[HTML]{F6F7D9} 
\textbf{seating}            & 69              & 69            & 0                       \\ \hline
\rowcolor[HTML]{F8DFDF} 
\textbf{picture}            & 27              & \textbf{30}    & -10                     \\ \hline
\rowcolor[HTML]{F8DFDF} 
\textbf{cushion}            & 46              & \textbf{56}   & -17.86                  \\ \hline
\rowcolor[HTML]{F8DFDF} 
\textbf{towel}              & 12              & \textbf{15}   & -20                     \\ \hline
\textbf{Average}            & \textbf{37} $\pm 3$   & $35 \pm 3$ & 6.55                   \\ \hline
\end{tabular}
}

\label{tab:ObjIDResults}
\end{table}

\subsection{Performance on the Habitat Challenge}
\label{Habitat}
We evaluate \stubborn on the 2021 Habitat Object Navigation Challenge with the test-standard split. Results in Table \ref{tab:Habitat} show that our agent (with code name $yummi\_the\_magic\_cat$) has the second best performance in SPL (trailing by a very small margin) and Success Rate. These results convincingly show that our method that doesnot uses semantics for exploration is a strong baseline for the \objnav task. 

\begin{table}[]
\caption{\small  We report the leaderboard entries on standard test split of the \textbf{2021 Habitat Object Navigation Challenge} Entries with the superscript 2020 and 2021 indicate challenge winners in the corresponding year. Our Method (named Yuumi\_the\_magic\_cat) is ranked second by a minute margin based on SPL and accuracy. It substantially outperforms \semexp, the agent we have built upon.}
\centering
\resizebox{\columnwidth}{!}{%
\begin{tabular}{|l c c|}
\hline
Models                            & \textbf{SPL}    & \textbf{Success Rate \%} \\ \hline

(1) Habitat on Web (IL-HD)           & \textbf{0.099}  & \textbf{27.8}        \\
\rowcolor{LightCyan}
(2) yuumi\_the\_magic\_cat(Our Method) & 0.098  & 23.7        \\ 
(3) TreasureHunt\cite{maksymets2021thda}                     & 0.089  & 21.4        \\ 

(4) PONI (PF)                        & 0.088  & 20          \\ 
(5) EmbCLIP \cite{Khandelwal2021objectnav-clip}                       & 0.078  & 18.1        \\ 
\rowcolor{LightYellow}
(6) Arnold (\semexp)$^{2020}$ \cite{chaplot2020gose}                 & 0.0707 & 17.85       \\ 
(7) OVRL (RGBD)          & 0.076  & 23.2        \\ 
(8) Red Rabbit 6-Act Base (EEAux)$^{2021}$ \cite{ye2021auxiliary}          & 0.062  & 23.7        \\  \hline
\end{tabular}
}

\label{tab:Habitat}
\end{table}

\section{Related Work}
\label{sec:related-work}

\textbf{Object Navigation}
Reinforcement learning (RL) has been employed to tackle object navigation in several recent works~\cite{ye2021auxiliary,Wahid2020objectnav-sac,Khandelwal2021objectnav-clip,maksymets2021thda}. These methods take goal objects as the policy's inputs and reward the policy upon reaching goal. In addition, Mousavian et al. \cite{Mousavian2019VisualRF} use imitation learning to learn a policy matching the behaviors of the optimal path planner with privileged information of the environment. In contrast, other works build explicit spatial representations. Chaplot et al. \cite{chaplot2020gose} maintains a semantic map for predicting locations of goal objects. Gupta et al. \cite{Gupta2019CMP} construct a belief map of goal locations for planning. Yang et al. \cite{Yang2019scenepriors} construct an object-to-object knowledge graph and using a graph convolutional network to generate semantic priors for object locations.


\textbf{Object Detection} Recent works have drawn upon pre-trained computer vision models \cite{he2017mask,chen2017deeplabv3,jiang2018rednet}. As per-frame predictions could produce inconsistent results across time, prior works tried improving accuracy by temporal relationships. David et al. \cite{Nilsson2018SemVideoSegGRU} and Zhu et al. \cite{zhu17dff} use optical flow networks to propagate predictions over time.  Liu et al.~\cite{Liu2020EfficientSV} use temporal consistency loss during training. Wang et al.~\cite{Wang2021TemporalMA} employ attention to relate the current frame and previous frames. While we leave incorporating such techniques to future works, we try to improve temporal consistency via an additional learned model that refines the single-frame object detection system used in \cite{chaplot2020gose}.


\textbf{Exploration for Navigation} Exploration is a critical component for navigation tasks. Chen et al. \cite{chen2019learning} train an exploration policy by RL with coverage rewards. Kollar et al. \cite{kollar2008trajectory} uses trajectory optimization to derive an efficient exploration strategy. Pathak et al. \cite{pathak2017curiosity} train an exploration policy using prediction errors of the forward dynamics model. In contrast, we use a rule-based exploration strategy alternating among corners of the map.



\bibliographystyle{IEEEtran}
\bibliography{IEEEabrv,references.bib}

\begin{thebibliography}{10}
\providecommand{\url}[1]{#1}
\csname url@rmstyle\endcsname
\providecommand{\newblock}{\relax}
\providecommand{\bibinfo}[2]{#2}
\providecommand\BIBentrySTDinterwordspacing{\spaceskip=0pt\relax}
\providecommand\BIBentryALTinterwordstretchfactor{4}
\providecommand\BIBentryALTinterwordspacing{\spaceskip=\fontdimen2\font plus
\BIBentryALTinterwordstretchfactor\fontdimen3\font minus
  \fontdimen4\font\relax}
\providecommand\BIBforeignlanguage[2]{{%
\expandafter\ifx\csname l@#1\endcsname\relax
\typeout{** WARNING: IEEEtran.bst: No hyphenation pattern has been}%
\typeout{** loaded for the language `#1'. Using the pattern for}%
\typeout{** the default language instead.}%
\else
\language=\csname l@#1\endcsname
\fi
#2}}

\bibitem{savva2019habitat}
M.~Savva, A.~Kadian, O.~Maksymets, Y.~Zhao, E.~Wijmans, B.~Jain, J.~Straub,
  J.~Liu, V.~Koltun, J.~Malik, D.~Parikh, and D.~Batra, ``Habitat: A platform
  for embodied ai research,'' \emph{ICCV}, pp. 9338--9346, 2019.

\bibitem{Anderson2018OnEO}
P.~Anderson, A.~X. Chang, D.~S. Chaplot, A.~Dosovitskiy, S.~Gupta, V.~Koltun,
  J.~Kosecka, J.~Malik, R.~Mottaghi, M.~Savva, and A.~R. Zamir, ``On evaluation
  of embodied navigation agents,'' \emph{arXiv preprint arXiv:1807.06757},
  2018.

\bibitem{chaplot2020gose}
D.~S. Chaplot, D.~Gandhi, A.~Gupta, and R.~Salakhutdinov, ``Object goal
  navigation using goal-oriented semantic exploration,'' \emph{NeurIPS}, 2020.

\bibitem{ye2021auxiliary}
J.~Ye, D.~Batra, A.~Das, and E.~Wijmans, ``Auxiliary tasks and exploration
  enable object navigation,'' \emph{ICCV}, 2021.

\bibitem{dai2021coatnet}
Z.~Dai, H.~Liu, Q.~Le, and M.~Tan, ``Coatnet: Marrying convolution and
  attention for all data sizes,'' \emph{Advances in Neural Information
  Processing Systems}, vol.~34, 2021.

\bibitem{grisetti2007improved}
G.~Grisetti, C.~Stachniss, and W.~Burgard, ``Improved techniques for grid
  mapping with rao-blackwellized particle filters,'' \emph{IEEE transactions on
  Robotics}, vol.~23, no.~1, pp. 34--46, 2007.

\bibitem{grisettiyz2005improving}
G.~Grisettiyz, C.~Stachniss, and W.~Burgard, ``Improving grid-based slam with
  rao-blackwellized particle filters by adaptive proposals and selective
  resampling,'' in \emph{Proceedings of the 2005 IEEE international conference
  on robotics and automation}.\hskip 1em plus 0.5em minus 0.4em\relax IEEE,
  2005, pp. 2432--2437.

\bibitem{mur2015orb}
R.~Mur-Artal, J.~M.~M. Montiel, and J.~D. Tardos, ``Orb-slam: a versatile and
  accurate monocular slam system,'' \emph{IEEE transactions on robotics},
  vol.~31, no.~5, pp. 1147--1163, 2015.

\bibitem{keselman2017intel}
L.~Keselman, J.~Iselin~Woodfill, A.~Grunnet-Jepsen, and A.~Bhowmik, ``Intel
  realsense stereoscopic depth cameras,'' in \emph{Proceedings of the IEEE
  Conference on Computer Vision and Pattern Recognition Workshops}, 2017, pp.
  1--10.

\bibitem{batra2020objectnav}
D.~Batra, A.~Gokaslan, A.~Kembhavi, O.~Maksymets, R.~Mottaghi, M.~Savva,
  A.~Toshev, and E.~Wijmans, ``Object{N}av {R}evisited: {O}n {E}valuation of
  {E}mbodied {A}gents {N}avigating to {O}bjects,'' in \emph{arXiv:2006.13171},
  2020.

\bibitem{mp3d}
A.~Chang, A.~Dai, T.~Funkhouser, M.~Halber, M.~Niessner, M.~Savva, S.~Song,
  A.~Zeng, and Y.~Zhang, ``Matterport3d: Learning from rgb-d data in indoor
  environments,'' \emph{International Conference on 3D Vision (3DV)}, 2017,
  license available at http://kaldir.vc.in.tum.de/matterport/MP\_TOS.pdf.

\bibitem{efron1985bootstrap}
B.~Efron and R.~Tibshirani, ``The bootstrap method for assessing statistical
  accuracy,'' \emph{Behaviormetrika}, vol.~12, no.~17, pp. 1--35, 1985.

\bibitem{ffm}
J.~A. Sethian, ``A fast marching level set method for monotonically advancing
  fronts,'' \emph{Proceedings of the National Academy of Sciences}, vol.~93,
  no.~4, pp. 1591--1595, 1996.

\bibitem{zhu2012human}
M.-Q. Zhu, Z.-L. Wang, and Z.-H. Chen, ``Human visual intelligence and particle
  filter based robust object tracking algorithm,'' \emph{Control and Decision},
  vol.~27, no.~11, pp. 1720--1724, 2012.

\bibitem{jiang2018rednet}
J.~Jiang, L.~Zheng, F.~Luo, and Z.~Zhang, ``Rednet: Residual encoder-decoder
  network for indoor rgb-d semantic segmentation,'' \emph{arXiv preprint
  arXiv:1806.01054}, 2018.

\bibitem{frontier}
B.~Yamauchi, ``A frontier-based approach for autonomous exploration,'' in
  \emph{Proceedings 1997 IEEE International Symposium on Computational
  Intelligence in Robotics and Automation CIRA'97.'Towards New Computational
  Principles for Robotics and Automation'}.\hskip 1em plus 0.5em minus
  0.4em\relax IEEE, 1997, pp. 146--151.

\bibitem{maksymets2021thda}
O.~Maksymets, V.~Cartillier, A.~Gokaslan, E.~Wijmans, W.~Galuba, S.~Lee, and
  D.~Batra, ``Thda: Treasure hunt data augmentation for semantic navigation,''
  in \emph{Proceedings of the IEEE/CVF International Conference on Computer
  Vision}, 2021, pp. 15\,374--15\,383.

\bibitem{Khandelwal2021objectnav-clip}
A.~Khandelwal, L.~Weihs, R.~Mottaghi, and A.~Kembhavi, ``Simple but effective:
  Clip embeddings for embodied ai,'' \emph{arXiv preprint arXiv:2111.09888},
  2021.

\bibitem{Wahid2020objectnav-sac}
A.~Wahid, A.~Stone, K.~Chen, B.~Ichter, and A.~Toshev, ``Learning
  object-conditioned exploration using distributed soft actor critic,'' 2020.

\bibitem{Mousavian2019VisualRF}
A.~Mousavian, A.~Toshev, M.~Fiser, J.~Kosecka, and J.~Davidson, ``Visual
  representations for semantic target driven navigation,'' \emph{2019
  International Conference on Robotics and Automation (ICRA)}, pp. 8846--8852,
  2019.

\bibitem{Gupta2019CMP}
S.~Gupta, V.~Tolani, J.~Davidson, S.~Levine, R.~Sukthankar, and J.~Malik,
  ``Cognitive mapping and planning for visual navigation,'' \emph{International
  Journal of Computer Vision}, vol. 128, pp. 1311--1330, 2019.

\bibitem{Yang2019scenepriors}
W.~Yang, X.~Wang, A.~Farhadi, A.~K. Gupta, and R.~Mottaghi, ``Visual semantic
  navigation using scene priors,'' \emph{arXiv preprint arXiv:1810.06543},
  2019.

\bibitem{he2017mask}
K.~He, G.~Gkioxari, P.~Doll{\'a}r, and R.~Girshick, ``Mask r-cnn,'' in
  \emph{Proceedings of the IEEE international conference on computer vision},
  2017, pp. 2961--2969.

\bibitem{chen2017deeplabv3}
L.-C. Chen, G.~Papandreou, F.~Schroff, and H.~Adam, ``Rethinking atrous
  convolution for semantic image segmentation,'' \emph{arXiv preprint
  arXiv:1706.05587}, 2017.

\bibitem{Nilsson2018SemVideoSegGRU}
D.~Nilsson and C.~Sminchisescu, ``Semantic video segmentation by gated
  recurrent flow propagation,'' in \emph{CVPR}, 2018.

\bibitem{zhu17dff}
X.~Zhu, Y.~Xiong, J.~Dai, L.~Yuan, and Y.~Wei, ``Deep feature flow for video
  recognition,'' in \emph{CVPR}, 2017.

\bibitem{Liu2020EfficientSV}
Y.~Liu, C.~Shen, C.~Yu, and J.~Wang, ``Efficient semantic video segmentation
  with per-frame inference,'' in \emph{ECCV}, 2020.

\bibitem{Wang2021TemporalMA}
H.~Wang, W.~Wang, and J.~Liu, ``Temporal memory attention for video semantic
  segmentation,'' \emph{arXiv preprint arXiv:2102.08643}, 2021.

\bibitem{chen2019learning}
T.~Chen, S.~Gupta, and A.~Gupta, ``Learning exploration policies for
  navigation,'' \emph{arXiv preprint arXiv:1903.01959}, 2019.

\bibitem{kollar2008trajectory}
T.~Kollar and N.~Roy, ``Trajectory optimization using reinforcement learning
  for map exploration,'' \emph{The International Journal of Robotics Research},
  vol.~27, no.~2, pp. 175--196, 2008.

\bibitem{pathak2017curiosity}
D.~Pathak, P.~Agrawal, A.~A. Efros, and T.~Darrell, ``Curiosity-driven
  exploration by self-supervised prediction,'' in \emph{International
  conference on machine learning}.\hskip 1em plus 0.5em minus 0.4em\relax PMLR,
  2017, pp. 2778--2787.

\end{thebibliography}

\end{document}